\documentclass[letterpaper]{article} 
\usepackage{aaai24}  
\usepackage{times}  
\usepackage{helvet}  
\usepackage{courier}  
\usepackage[hyphens]{url}  
\usepackage{graphicx} 
\usepackage{natbib}  
\usepackage{caption} 
\usepackage{algorithm}
\usepackage{algorithmic}

\usepackage{newfloat}
\usepackage{listings}
\DeclareCaptionStyle{ruled}{labelfont=normalfont,labelsep=colon,strut=off} 
\floatstyle{ruled}
\newfloat{listing}{tb}{lst}{}
\floatname{listing}{Listing}
\usepackage{amsmath}    
\usepackage{amsfonts} 
\usepackage{multirow}
\usepackage{color}
\usepackage{graphicx}
\usepackage{makecell}
\usepackage{pifont}

\definecolor{pzy}{RGB}{0,0,255}
\newcommand{\eg}{{\emph{e.g.}}}

\newcommand{\ie}{{\emph{i.e.}}}

\title{Semi-Supervised Class-Agnostic Motion Prediction with Pseudo Label Regeneration and BEVMix}
\author{
    Kewei Wang\textsuperscript{\rm 1,2},
    Yizheng Wu\textsuperscript{\rm 1,2},
    Zhiyu Pan\textsuperscript{\rm 2},
    Xingyi Li\textsuperscript{\rm 1,2},
    Ke Xian\textsuperscript{\rm 1},
    Zhe Wang\textsuperscript{\rm 3},
    Zhiguo Cao\textsuperscript{\rm 2},
    Guosheng Lin\textsuperscript{\rm 1}\thanks{Corresponding author.}
}
\affiliations{
    \textsuperscript{\rm 1} S-Lab, Nanyang Technological University\\
    \textsuperscript{\rm 2} Key Laboratory of Image Processing and Intelligent Control, Ministry of Education; School of Artificial Intelligence and Automation, Huazhong University of Science and Technology, Wuhan 430074, China\\
    \textsuperscript{\rm 3} SenseTime Research\\
    \{kewei.wang, gslin\}@ntu.edu.sg, \{wangkewei, zgcao\}@hust.edu.cn

}

\usepackage{bibentry}

\begin{document}

\maketitle
\begin{abstract}
Class-agnostic motion prediction methods aim to comprehend motion within open-world scenarios, holding significance for autonomous driving systems.
However, training a high-performance model in a fully-supervised manner always requires substantial amounts of manually annotated data, which can be both expensive and time-consuming to obtain.
To address this challenge, our study explores the potential of semi-supervised learning (SSL) for class-agnostic motion prediction. Our SSL framework adopts a consistency-based self-training paradigm, enabling the model to learn from unlabeled data by generating pseudo labels through test-time inference. To improve the quality of pseudo labels, we propose a novel motion selection and re-generation module. This module effectively selects reliable pseudo labels and re-generates unreliable ones.
Furthermore, we propose two data augmentation strategies: temporal sampling and BEVMix. These strategies facilitate consistency regularization in SSL.
Experiments conducted on nuScenes demonstrate that our SSL method can surpass the self-supervised approach by a large margin by utilizing only a tiny fraction of labeled data. Furthermore, our method exhibits comparable performance to  weakly and some fully supervised methods. These results highlight the ability of our method to strike a favorable balance between annotation costs and performance. 
Code will be available at \tt{https://github.com/kwwcv/SSMP}.

\end{abstract}
\section{Introduction}
Understanding the motion behavior within dynamic environments is crucial for a variety of autonomous systems. Traditional methods approach motion perception through trajectory prediction~\cite{chai2019multipath, chang2019argoverse,fang2020tpnet, liang2020pnpnet}. However, these approaches may face challenges when handling categories that have not been seen in the training set, mainly due to their reliance on object detection~\cite{Wu2020MotionNetJP}.
To address this challenge, class-agnostic motion prediction task~\cite{schreiber2019long,Wu2020MotionNetJP,Wang2022BESTISI, Wei2022SpatiotemporalTA} is proposed to provide complementary information.
These methods take a sequence of previous point clouds as input and predict the future displacements for each Bird's Eye View (BEV) cell.
Although fully-supervised methods have achieved significant success, they often require substantial amounts of annotated point cloud data, which can be both costly and time-consuming to acquire. 
To overcome this limitation, self-supervised and weakly supervised approaches have been proposed~\cite{Luo2021SelfSupervisedPM, li2023weakly}. However, noticeable performance gaps still exist between the best results obtained from these annotation-efficient methods and those achieved by fully-supervised methods. These annotation-efficient methods generate supervision by matching source points with target points, in which handling fast motions becomes challenging~\cite{li2023weakly}. Therefore, we delve into exploiting the use of limited, accurate motion labels to generate supervision for easily accessible unlabeled data, \ie, the semi-supervised motion prediction.

Semi-supervised learning (SSL) seeks to largely alleviate the need for labeled data by leveraging unlabeled data~\cite{mixmatch, meanteacher, sohn2020fixmatch,STAC,instanceTeacher,SoftTeacher}. In this study, we adopt one of the most widely used SSL paradigms, consistency-based self-training~\cite{sohn2020fixmatch}. 
The key idea is to first generate reliable pseudo labels for the unlabeled data (pseudo-labeling), and then train the model to predict the pseudo label when feeding the unlabeled data with perturbation (consistency regularization). Specifically, for the motion prediction task, we specially design two novel data augmentations for consistency regularization and the motion select and re-generate module (MSRM) for pseudo-labeling.

Pseudo-labeling~\cite{lee2013pseudo} plays a role in enhancing semi-supervised learning performance by selecting reliable pseudo-labels while discarding unreliable ones. However, unlike in image classification and object detection, there is currently no appropriate metric like the confidence score to evaluate the reliability of predicted motion. Given that motion represents the displacement from the current cell to the corresponding future cell, an accurate motion would ideally lead the current cell to overlap precisely with the corresponding cell after warping. Based on this fact, we design the motion select and re-generate module (MSRM) to select reliable predicted motions by measuring the distance between the warped cell and its correspondence. Additionally, labels of the discarded cells are re-generated by the neighbor reliable labels based on the assumption of local smoothness.

Two simple yet efficient data augmentations for motion prediction are proposed as Temporal-Sampling (TS) and BEVMix. These augmentations will serve as strong augmentations to drive the weak-to-strong consistency regularization. Specifically, TS is used to sub-sample the input sequence temporally to generate additional artifact samples with larger motion labels. Meanwhile, BEVMix is designed to mix two different BEV sequences to synthesize new training samples with a large diversity. This diversity can enhance the generalization ability of the model, and therefore improve the prediction performance. 

Following previous works~\cite{Wu2020MotionNetJP, Luo2021SelfSupervisedPM}, we test the efficacy of our method on a large-scale autonomous driving dataset, nuScenes~\cite{Caesar2019nuScenesAM}. We use 1\%,5\%, and 10\% of labeled data as labeled sets and the remainder as unlabeled sets to evaluate the effectiveness of our semi-supervised method.  

Overall, the contributions of this paper are as follows:
\begin{itemize}
    \item We are the first to explore semi-supervised learning in the class-agnostic motion prediction task.
    
    \item We propose MSRM to filter unreliable pseudo labels and re-generate them from reliable ones, which enables the model to learn more from high-quality pseudo labels. 
    
    \item We introduce two new augmentations, temporal-sampling and BEVMix, for motion prediction, which facilitate consistency regularization in SSL.
\end{itemize}
\section{Related Work}
\textbf{Class-Agnostic Motion Prediction.}
Motion prediction aims to predict the future motion of agents based on past observations. Traditional approaches achieve motion prediction through object detection~\cite{Zhou2017VoxelNetEL, lang2019pointpillars}, followed by subsequent trajectory prediction~\cite{chai2019multipath, chang2019argoverse, djuric2020uncertainty,fang2020tpnet}. Relying on object detection, however, these approaches may fail to handle unknown object classes~\cite{Wu2020MotionNetJP}. 
To provide complementary motion information, class-agnostic motion prediction methods avoid dependence on detection and predict motion directly. These methods represent the environment with BEV maps derived from point clouds and aim to predict the 2D displacement vector for each BEV cell along the horizontal plane. MotionNet~\cite{Wu2020MotionNetJP} and BE-STI~\cite{Wei2022SpatiotemporalTA} proposes to perform joint category perception and motion prediction from the BEV maps. LSTM-ED~\cite{schreiber2019long} introduce convolutional LSTM~\cite{Shi2015ConvolutionalLN} to aggregate temporal context. 
Recently, annotation-efficiency methods such as PillarMotion~\cite{Luo2021SelfSupervisedPM} and WeakMotioinNet~\cite{li2023weakly} have been proposed to train motion prediction models in a self-supervised and weakly-supervised manner, respectively. While ~\cite{li2023weakly} aims to utilize easier-acquired annotations, we explore to make trade-off between performance and the quantity of annotations. \\
\textbf{Scene Flow Estimation.} 
Scene flow estimation methods~\cite{Liu2018FlowNet3DLS,Gu2019HPLFlowNetHP, sun2018pwc,behl2019pointflownet, wang2021hierarchical} produce 3D motion fields in a dense manner. In comparison with class-agnostic motion prediction methods that seek to predict the displacements to future from past observations, scene flow estimation aims to estimate the motion flow  between two observed point clouds. Nonetheless, estimating dense 3D flow always demands large computation, making it unfeasible for real-time autonomous systems~\cite{li2023weakly}. Moreover, the direct application of scene flow estimation to actual LiDAR point clouds presents inherent challenges, primarily attributed to the lack of consistent one-to-one correspondences~\cite{Wang2022BESTISI}.
\\
\textbf{Semi-Supervised Learning (SSL).} SSL integrates information from limited labeled data and extensive unlabeled data. Consistency-based regularization~\cite{mixmatch,xie2020unsupervised,laine2016temporal,berthelot2019remixmatch,meanteacher} applies a consistency loss by enforcing invariance on unlabeled data under different augmentations. Pseudo-labeling relies on the model's high confident predictions to produce pseudo-labels~\cite{lee2013pseudo,bachman2014learning,arazo2020pseudo} for unlabeled data and trains them jointly with labeled data. FixMatch~\cite{sohn2020fixmatch} is a combination of both consistency-based regularization and pseudo-labeling approaches. In this study, we adopt the consistency-based self-training paradigm from FixMatch, enhancing the performance of semi-supervised motion prediction through both consistency regularization and pseudo-labeling aspects.
\begin{figure*}[t]
	\begin{center}
		\includegraphics[width=1.0\linewidth]{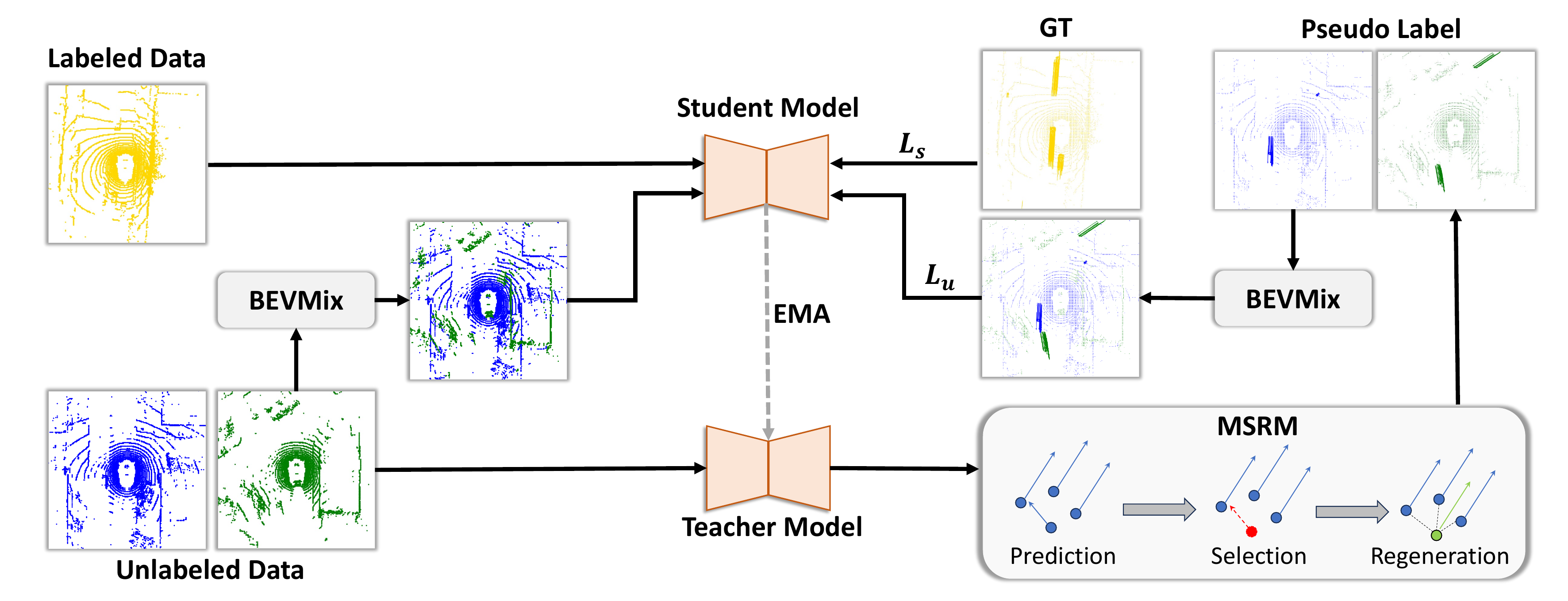}
		\caption{Overview of the proposed approach. Unlabeled samples are first fed to the teacher model to produce pseudo labels, followed by MSRM to improve the quality. Subsequently, the unlabeled samples are mixed using BEVMix and fed to the student model to compute the unlabeled loss. Concurrently, labeled samples are also fed to the student model to compute the supervised loss. The weights of the teacher model are updated from the weights of the student model by Exponential Moving Average (EMA) in every iteration.}
		\label{fig:flowchart}
	\end{center}
\end{figure*}

\section{Semi-Supervised Motion Prediction}
\subsection{Problem Formulation}
Class-agnostic motion prediction methods take a temporal sequence of LiDAR point cloud frames as input, where all the point clouds are synchronized to the current coordinate system~\cite{Wu2020MotionNetJP}. We denote synchronized point cloud captured at time $t$ as $P^t$. $P^t$ is then discretized into dense voxels $V^t\in \{0,1\}^{H\times W \times C}$, where 0 indicates the voxel is empty, 1 indicates the voxel is occupied by at least 1 point, and $H$, $W$, $C$ are the voxel numbers along $X$, $Y$, $Z$ axis respectively.
We view $C$ as the feature dimension of an image and $V^t$ as a virtual BEV map with $H\times W$ cells. Then the motion field $M \in \mathbb{R}^{H\times W\times 2}$ in the BEV map is defined as the 2D displacement of each BEV cell to the next timestamp. 
Taking BEV map sequence $\mathcal{V}=\{V^t\}_{t=1}^T$ as input, the motion prediction model aims to predict the motion field $M$.
For SSL, our goal is to train a motion prediction model by leveraging both a large amount of unlabeled data $\mathcal{D}_u={(\mathcal{V}^u)}$ and a smaller set of labeled data $\mathcal{D}_l={(\mathcal{V}^l,M^l)}$.

\subsection{SSL for Motion Prediction}
For semi-supervised learning, we basically adopt the framework of mean-teacher~\cite{meanteacher} for its effectiveness and flexibility. This involves two stages of training, where in the first stage, we train a teacher model using all available labeled data, and in the second stage, we train a student model using both labeled and unlabeled data with strong-to-weak consistency. The steps of our semi-supervised training are summarized as follows:
\begin{enumerate}
    \item Train a teacher model on labeled samples.
    \item Generate pseudo labels of unlabeled samples using the trained teacher model and weak augmentations.
    \item Select reliable pseudo labels and re-generate the unreliable ones.
    \item Apply strong data augmentations to unlabeled samples and train a student model with both the labeled and unlabeled samples.
    \item Update teacher model with student model. Repeat steps 2-4.
\end{enumerate}

\subsection{Training Teacher Model} There are two branches of our pipeline, one teacher model $\mathcal{G}_{\theta^t}$ and one student model $\mathcal{G}_{\theta^s}$. We first train the teacher model on the labeled dataset with supervised loss:
\begin{equation}
    \mathcal{L}_s =\ell_{smooth_{l1}}\big(\mathcal{G}_\theta(\mathcal{V}^l),M^l\big).
\end{equation}

\subsection{Motion Select and Re-generate Module (MSRM)}
\begin{figure}[t]
	\begin{center}
		\includegraphics[width=1.0\linewidth,]{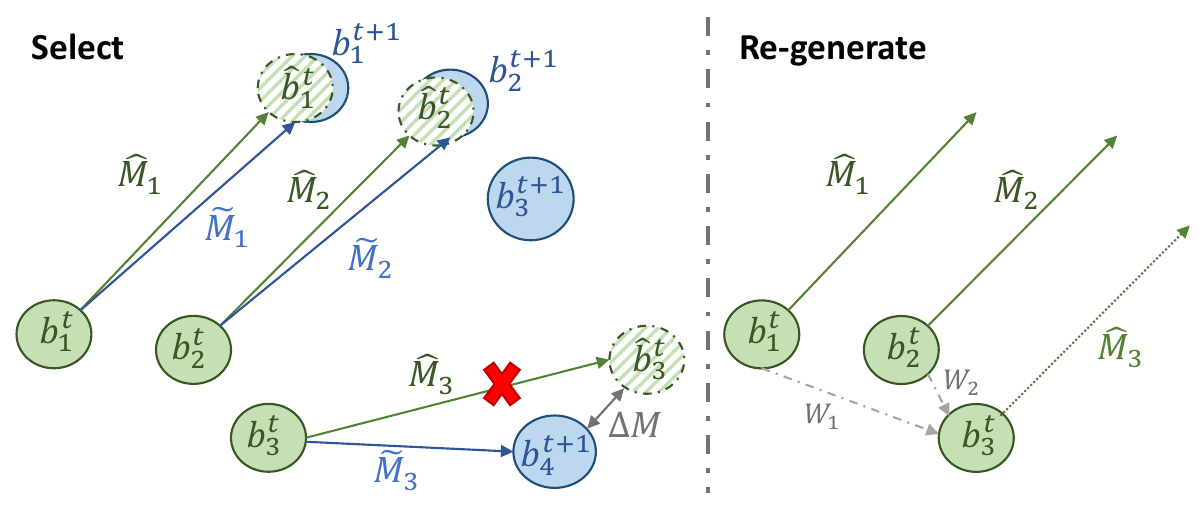}
		\caption{Diagram of MSRM. Select: In the case of an inaccurate pseudo label $\hat{M}_3$, where $b^t_3$ is incorrectly matched with a noise $b^{t+1}_4$, resulting in a large $\Delta M$. Re-generate: the pseudo label of $b^t_3$ is then re-generated by the weighted mean of neighbor reliable labels.}
		\label{fig:msrm}
	\end{center}
\end{figure}

We first perform test-time inference using the trained teacher model on unlabeled data to generate pseudo labels $\hat{M}^u$. \\
\noindent\textbf{Challenge in selecting reliable pseudo labels.}
However, trained on limited labeled data, the teacher model produces low-quality pseudo labels. Learning from these low-quality labels will negatively impact the performance of the model. For image classification and object detection, reliable (high-quality) pseudo labels can be selected by classification score or objectiveness score. But for motion prediction, as a regression task, there is no explicit metric to judge the reliability. To this end, we propose the motion select and re-generate module (MSRM), which shows effectiveness in generating reliable pseudo labels for the motion prediction task.\\
\noindent\textbf{Approach to evaluate the reliability of motion labels.}
As the pseudo motion labels $\hat{M}^u$ indicate the displacement of each BEV cell from the current frame to the future one, if the motion is accurate, the warped BEV cells would be exactly overlapped with the corresponding future ones. Based on this, we can first warp the current BEV cells with the pseudo label $\hat{M}^u$, and then find the corresponding cells in the future frame. Finally, we can select the reliable pseudo label based on the distance between the corresponding ones.
The correspondence can be found by solving the optimal transport problem. We define $B^t=\{b^t_i \in \mathbb{R}^2 \}^{N_t}_{i=1}$ as the 2D coordinates of BEV cells on $V^t$, where $N_t$ is the number of non-empty cells. The warped coordinates are denoted as:
\begin{equation}
    \hat{B}^t = B^t + \hat{M}^u
    \label{eq:warp}
\end{equation}
Ideally, $\hat{B}^t$ can be matched with $B^{t+1}=\{b^{t+1}_j \in \mathbb{R}^2 \}^{N_{t+1}}_{j=1}$ with matching matrix $\pi \in \{0,1\}^{N_t\times N_{t+1}}$:
\begin{equation}
    \hat{B}^t = \pi B^{t+1},
\label{eq:permute}
\end{equation}
To find the optimal matching matrix, we first compute the cost matrix $C$ by the pairwise distances:
\begin{equation}
    C_{ij} = 1-\exp{(-\frac{||\hat{b}_i^t-b_j^{t+1}||^2}{\theta_c})},
\end{equation}
where $\theta_c$ is a temperature parameter.
The optimal matching matrix can be approximated by solving an optimal transport problem~\cite{cuturi2013sinkhorn}:
\begin{equation}
\begin{aligned}
    &\pi^* = \mathop{\arg\min}\limits_{\pi} \sum_{i,j}C_{ij}\pi_{ij}\\
    &s.t.\quad \pi \mathbf{1}_{N_{t+1}} = \frac{1}{N_t} \mathbf{1}_{N_t}, \pi^T \mathbf{1}_{N_t} = \frac{1}{N_{t+1}} \mathbf{1}_{N_{t+1}},
\end{aligned}
\end{equation}
where $\pi^*$ is the optimal matching matrix. 
According to Eq.~\ref{eq:warp} and Eq.~\ref{eq:permute}, we can obtained auxiliary pseudo labels:
\begin{equation}
    \tilde{M}^u = \pi^* B^{t+1} - B^t.
\end{equation}
Finally, we can evaluate the motion pseudo label's reliability by the difference between $\hat{M}^u$ and $\tilde{M}^u$. The difference is computed by:
\begin{equation}
    \Delta M = ||\hat{M}^u-\tilde{M}^u||_2.
\end{equation}
\\
\begin{algorithm}
	\caption{Pseudo Label Re-generation}
     \textbf{Input}: $I_R$, $I_{UR}$, $B$\\
     \textbf{Output}: $I^u$, $M^u$
	\label{alg1}
	\begin{algorithmic}[1]
        \STATE $I^u \leftarrow I_R $, $M^u\leftarrow \hat{M}^u(I_R)$ 
        \FOR{$i\in I_{UR}$}
        \STATE $D$ $\leftarrow$ $||B(i)-B(I_R)||_2$ 
        \STATE \texttt{// $I_K$: index of the K neighbors}
        \STATE $D_K$, $I_K$ $\leftarrow$ {\texttt{TopK}}($D$)  
        \STATE $D_K$, $I_K$ $\leftarrow$ $D_K(D_K < \beta)$, $I_K(D_K < \beta)$
        \STATE $M_K$ $\leftarrow$ $M^u(I_K)$

        \IF{\texttt{Len}($I_K$) is $zero$}
        \STATE   \texttt{continue}
        \ENDIF

        \STATE \texttt{// $\theta_w$: temperature parameter}
        \STATE $W$ $\leftarrow$ \texttt{exp}($-\frac{D_K}{\theta_w}$)
        \STATE $M_{mean}$ $\leftarrow$ \texttt{WeightedMean}($M_K, W$)
        
        \STATE $M_{dif}$ $\leftarrow$ \texttt{abs}($\frac{M_K-M_{mean}}{M_{mean}}$)
        
        \STATE $H$ $\leftarrow$ \texttt{exp} $\left(-\texttt{WeightedMean}(M_{dif}, W)\right)$

        \IF{$H > \gamma$}
        \STATE   $I^u$.\texttt{add}(i), $M^u$.\texttt{add}($M_{mean}$)
        \ENDIF
        
        \ENDFOR
	\end{algorithmic}  
\label{alg:PLR}
\end{algorithm}
\\
\textbf{Select and re-generate pseudo labels.}
Based on the difference $\Delta M$, we can obtain the indexes for cells with reliable label $I_R=\{i|\Delta M(i)<\mu\}$ and indexes for unreliable ones  $I_{UR}=\{i|\Delta M(i)\geq\mu\}$. The key idea is that if the pseudo label is accurate, the cost between the warped cell and its corresponding target cell will be small and the optimal solver is more likely to find the correct correspondence. The process is shown in Fig.~\ref{fig:msrm}.

Furthermore, inspired by local rigid characteristics of most objects (\eg, cars) in the driving scenario, we re-generate pseudo labels for unreliable cells from reliable adjacent labels, as shown in Algorithm~\ref{alg:PLR}. For each unreliable cell, we first find $K$ nearest reliable neighbor cells based on Euclidean distance (lines 3-5). And the neighbors are valid only if the distances are within the distance threshold $\beta$ (lines 6-7). Then we use these reliable labels to re-generate labels for unreliable cells. We first calculate the weights $W$ based on the distance, the closer the distance is, the larger the weight is (line 12). Then we calculate the weighted mean of the reliable neighbors as the re-generated labels (line 13). To evaluate the quality of the re-generated labels, We calculate the weighted differences between the K neighbors and $M_{mean}$ to measure the local consistency $H$  (lines 14-15). The idea is derived from the fact that most objects in the scene are almost rigid and each part of a single one should have similar motion. A larger $H$ indicates that the unreliable cell and its K neighbors are more likely from the same object. By discarding re-generated labels with low consistency $H$, we obtain the final pseudo labels $M^u$ and correspondent index $I^u$ for the unlabeled sample (line 17). We only compute unsupervised loss for cells within index $I^u$.

\subsection{Data Augmentations}
As discussed by FixMatch~\cite{sohn2020fixmatch}, weak-to-strong consistency is important for the performance of semi-supervised learning. 
In our framework, we establish consistency regularization through following process: firstly, the teacher model processes weakly augmented unlabeled samples, generating pseudo-labels through test-time inference. Subsequently, the same samples undergo strong augmentation and are fed to the student model for predictions. Finally, the regularization is achieved by enforcing consistency between the predictions from strongly augmented samples and the pseudo-labels derived from weakly augmented samples.
While data augmentation has been widely studied in image classification and object detection, it have not yet been thoroughly explored in the class-agnostic motion prediction task. To this end, we introduce two simple yet effective data augmentations called  temporal-sampling and BEVMix.\\
\noindent\textbf{Temporal-sampling.} Taking input $\mathcal{V}=\{V^t\}_{t=1}^T$, we can generate artificial sequence $\mathcal{V_{TS}}=\{V^1,V^3,V^5,...,V^t\}$ by uniform-sampling along temporal dimension. And correspondent motion ground truth will be $M_{TS}=2M$. We then pad $\mathcal{V_{TS}}=\{V^1,...,V^1,V^3,V^5,...,V^t\}$ with $V^1$ to make sure the sequence length is equal to $\mathcal{V}$. From temporal sampling, we simulate a scenario from original input $\mathcal{V}$, in which objects are static at the beginning and suddenly start to move with double speed.
By temporal-sampling, we can compensate for a number of fast-speed samples, which are relatively fewer in comparison to the static and slow ones in the original dataset. \\
\noindent \textbf{BEVMix.} Mix data augmentations aim to generate artifact samples by combining two existing samples.
Mixup~\cite{mixup} proposes to enforce smooth label transitions by linearly interpolating between pairs of input samples and their corresponding class labels. 
Instead of generating unnatural mixed samples like Mixup, CutMix~\cite{cutmix} proposes directly replacing the image region with a patch from another training image. 
These strategies are also lifted from 2D to 3D for point cloud recognition~\cite{PointMixup, pointcutmix}.
%
To be not limited to single objects, Mix3D~\cite{mix3d} proposes to mix two full point cloud scenes directly by combining all the points from two scenes as a single scene for point cloud segmentation. LaserMix~\cite{lasermix} leverages the spatial prior of point clouds and mix two scenes by LiDAR scans.
Although effective, previous data mix approaches mainly focus on single input while the motion prediction methods take sequence input, as it is important to observe the locations of each object throughout the temporal window to make accurate motion prediction. Directly applying data mix strategies like Cutmix may lead to a case in which we cannot find correspondent cells of a moving object in the current frame from previous frames. Additionally, these data mix approaches are mainly designed for classification or segmentation based on some task-specific priors, which are also not suitable for the motion prediction task.

To this end, we propose a simple but effective BEVMix to mix BEV sequences for class-agnostic motion prediction. As depicted in Algorithm~\ref{alg:bevmix}, we view one sample as foreground and the other as background. We first remove ground points from the "foreground" sample $\mathcal{V}^f$ by a ground segmentation algorithm~\cite{lee2022patchwork++} to reduce noise (line 2), obtain the coordinates of non-empty cells $\{B^f_t\}_{t=1}^T$, and then occupy "background" BEV map with $B^f$ (line 5). Meanwhile, the motion pseudo-label map of the background sample is also occupied by $M^f$ (line 7). Finally, we obtain the mixed sample $\{V^{mix}_t\}$ and the correspondent pseudo labels $M^{mix}$. 
The effect of the BEVMix is two-fold. For the objects in the "foreground" sample, BEVMix provides more scenes from the "background" sample. By occupying the "background" sample with the "foreground" sample, the temporal correspondent points in the "foreground" are maintained, which is crucial for motion prediction. For the "background" sample, BEVMix acts like sparse CutMix that replaces original cells with ones from the "foreground" sample. And thanks to the ground-removing and sparse distribution of non-empty cells in the BEV map, most of the temporal correspondent points in the "background" can be maintained after BEVMix, which is also beneficial.
\subsection{Training Student Model}
The unsupervised loss is computed by
\begin{equation}
    \mathcal{L}_u =\ell_{smooth_{l1}}\big(\mathcal{G}_{\theta^s}(\mathcal{V}^{mix}),M^{mix}\big).
\end{equation}
The overall loss for the student model is:
\begin{equation}
    \mathcal{L} = \mathcal{L}_s + \mathcal{L}_u.
\end{equation}
In every iteration, we use the Exponential Moving Average (EMA) to update the weights of the teacher model from the student model:
\begin{equation}
    \theta^t = \alpha \theta^t + (1-\alpha) \theta^s
\end{equation}

\begin{algorithm}
	\caption{BEVMix}
    \textbf{Input}:$\{B^f_t\}_{t=1}^T$, $\{B^b_t\}_{t=1}^T$, $M^f$, $M^b$, $\{V^f_t\}_{t=1}^T$, $\{V^b_t\}_{t=1}^T$\\
    \textbf{Output}:$\{V^{mix}_t\}$, $M^{mix}$
	\begin{algorithmic}[1]
        \STATE $\{B^{mix}_t\} \leftarrow \{B^b_t\}$, $M^{mix} \leftarrow M^b$, $\{V^{mix}_t\} \leftarrow \{V^b_t\}$
        
        \STATE $\{B^f_t\}_{t=1}^T \leftarrow$ \texttt{GroundRemove} ($\{B^f_t\}_{t=1}^T$)
        
        \FOR{$t\in [1,T]$}
            \FOR{$b^f \in B^f_t$}
            \STATE $V^{mix}_t (b^f) \leftarrow V^f_t (b^f)$
            
            \IF{$t=T$}
            \STATE $M^{mix}(b^f) \leftarrow M^l (b^f)$
            \ENDIF
            
            \ENDFOR
        \ENDFOR
	\end{algorithmic}  
\label{alg:bevmix}
\end{algorithm}
\section{Experiments}

\begin{table*}[t]
\centering
\renewcommand\tabcolsep{10pt}
\renewcommand\arraystretch{1.1}
\resizebox{\linewidth}{!}{
\begin{tabular}{l|c|cccccc}
\hline
\multirow{2}*{Method} &\multirow{2}*{Supervision}&\multicolumn{2}{c}{Static }&\multicolumn{2}{c}{Speed $\leq$ 5m/s (Slow)}&\multicolumn{2}{c}{Speed $\geq$ 5m/s (Fast)} \\
\cline{3-8}
&&Mean$\downarrow$ &Median$\downarrow$ &Mean$\downarrow$ &Median$\downarrow$ &Mean$\downarrow$ &Median$\downarrow$\\
\hline
LSTM-ED &Full. &0.0358 &0 &0.3551 &0.1044 &1.5885 &1.0003\\
MotionNet &Full. &0.0256 &0 &0.2565 &0.0962 &1.0744 &0.7332\\
BE-STI &Full. &0.0220 &0 &0.2115 &0.0929 &0.7511 &0.5413\\
PillarMotion &Self.& 0.1620 & 0.0010 &0.6972 &0.1758 &3.5504 &2.0844\\
WeakMotionNet &Weak.& 0.0243 & 0 &0.3316 &0.1201 &1.6422 &1.0319\\
\hline
\hline
\multirow{6}*{Ours} &Sup-only (1\%) &0.0173 &0 &0.4882 &0.1057 &3.7242 &2.5059\\
&Semi.(1\%) &\textbf{0.0153} &\textbf{0} &\textbf{0.3497} &\textbf{0.1020} &\textbf{1.9407} &\textbf{1.2173}\\
\cline{2-8}
&Sup-only(5\%) &0.0198 &0 &0.3532 &\textbf{0.0990} &2.6661 &1.6253\\
&Semi.(5\%) &\textbf{0.0183} &\textbf{0} &\textbf{0.3021} &0.0991 &\textbf{1.4516} &\textbf{0.8864}\\
\cline{2-8}
&Sup-only (10\%) &0.0245 &0 &0.3159 &0.0992 &1.9097 &1.1771\\
&Semi. (10\%) &\textbf{0.0218} &\textbf{0} &\textbf{0.2746} &\textbf{0.0996} &\textbf{1.2030} &\textbf{0.7880}\\
\hline
\end{tabular}
}
\caption{Results of motion prediction methods on the nuScenes dataset. "Full.", "Self.", "Weak.", and "Semi.", refer to fully-supervised, self-supervised, weakly-supervised, and semi-supervised training, respectively.}
\label{tab:main_results}
\end{table*}

\begin{figure*}[t]
	\begin{center}
		\includegraphics[width=1.0\linewidth, height=0.46\linewidth]{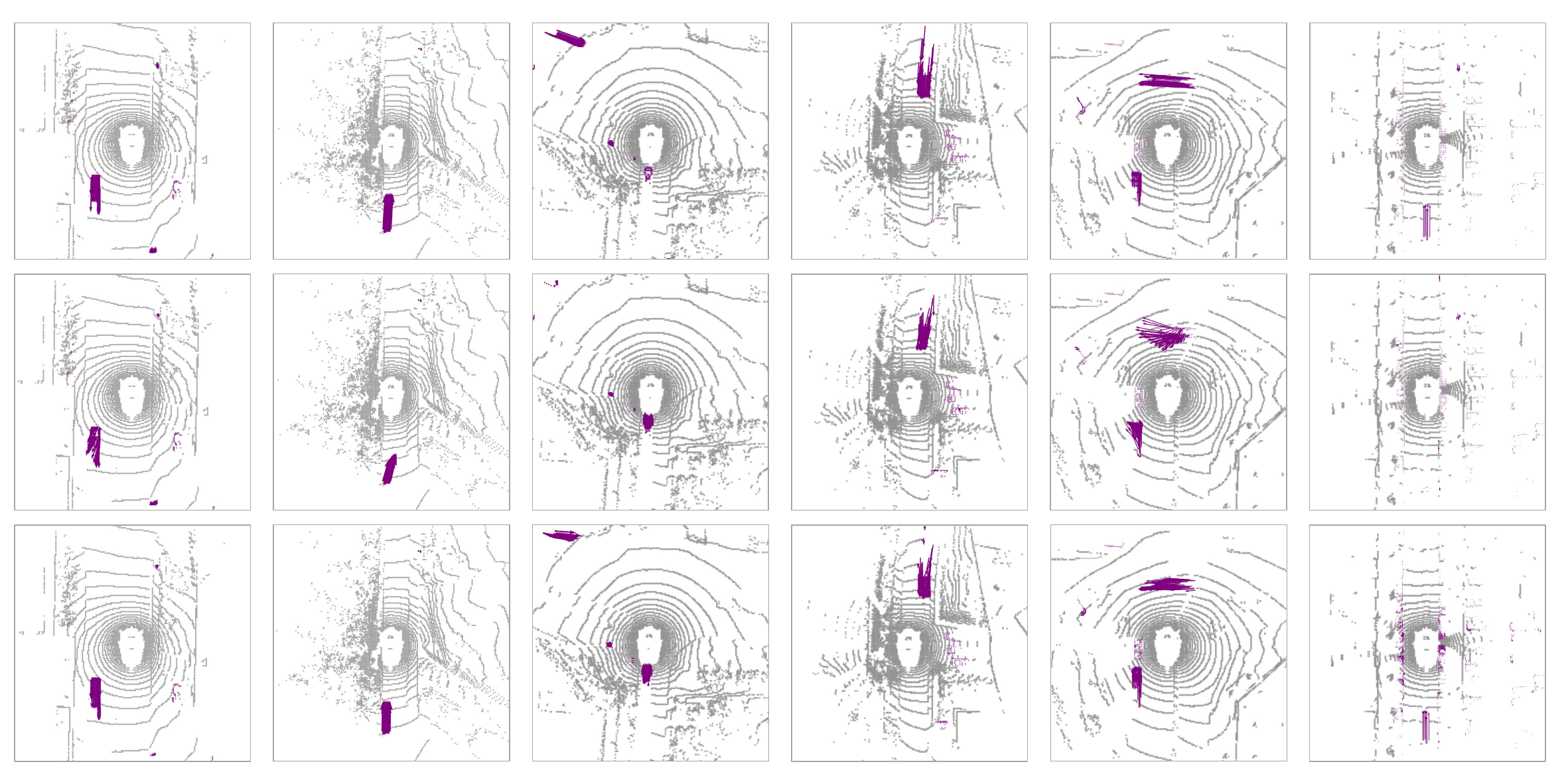}
		\caption{Qualitative results on the nuScenes dataset. First row: ground truth. Second row: results of training with 1\% labeled data. Third row: results of our semi-supervised method trained with 1\% labeled data and 99\% unlabeled data.}
		\label{fig:qualitative}
	\end{center}
\end{figure*}

\subsection{Settings} 
\textbf{Dataset.} Following previous work~\cite{Wu2020MotionNetJP, Luo2021SelfSupervisedPM, Wang2022BESTISI}, We evaluate the proposed approach on the large-scale self-driving dataset: nuScenes~\cite{Caesar2019nuScenesAM}, which contains 850 scenes with annotations. For fair comparisons, we follow MotionNet to use 500 scenes for training, 100 scenes for validation, and 250 scenes for testing.\\
\textbf{SSL settings.} For SSL settings, we random sample 1\%, 5\%, and 10\% training scenes as the labeled set and use the rest of the training scenes as an unlabeled set to simulate the scenario that large amounts of data are corrected, but only a minority of it has been annotated.\\
\textbf{Implementation details.} We follow the same pre-process with MotionNet, where the point clouds are cropped in the range of [-32m, 32m]$\times$ [-32m, 32m]$\times$ [-3m, 2m] and the voxel size is set to 0.25m $\times $0.25m $\times$ 0.4m along XYZ axis. We take the current sweep and the past four sweeps as input ($t=5$) and transform the past sweeps into the current coordinate system through ego-motion. Following previous works~\cite{Luo2021SelfSupervisedPM, li2023weakly}, we apply MotionNet as the baseline model. Adam~\cite{adam} is used as the optimizer. We implement our model in Pytorch~\cite{paszke2019pytorch} with a single A6000 GPU. 
We use the teacher model for evaluation following previous SSL works.
We use random flip as weak augmentations and proposed temporal sampling and BEVMix as strong augmentations.
The parameters $K$, $\mu$, $\beta$, $\gamma$, $\theta_c$, $\theta_w$, and $\alpha$ are set to 5, 1, 10, 0.6, 3, 5, and 0.999, respectively.
\\
\textbf{Evaluation metrics.} Following MotionNet, we evaluate the mean and median errors at different speed levels by dividing the grid cells into 3 groups according to the ground truth speeds: static, slow ($\leq$ 5m/s), and fast ($\geq$ 5m/s). Errors are measured by $L_2$ distances between the predicted displacements and the ground truth displacements for the next 1s. 
\begin{table}[t]
\centering
\renewcommand\tabcolsep{10pt}
\renewcommand\arraystretch{1.5}
\resizebox{\linewidth}{!}{
\begin{tabular}{cccc|ccc}
\hline
\multirow{2}*{MT} & \multirow{2}*{TS} &\multirow{2}*{MSRM} & \multirow{2}*{BEVMix} &Static &Slow &Fast \\
\cline{5-7}
&&&& \multicolumn{3}{c}{Mean Errors $\downarrow$}\\
\hline
\checkmark &  & &  &0.0133 &0.4218 &3.0868\\
\checkmark &\checkmark & & &0.0162 &0.4594 &2.9037\\
\checkmark &\checkmark &\checkmark & &0.0161 &0.3946 &2.7705 \\
\checkmark &\checkmark & &\checkmark &\textbf{0.0139} &0.3581 &2.1427 \\
\checkmark &\checkmark &\checkmark &\checkmark &0.0153 &\textbf{0.3497} &\textbf{1.9407} \\
\hline
\end{tabular}
}
\caption{Effectiveness of Temporal-sampling, MSRM, and BEVMix on nuScenes dataset in 1\% labeled data setting.}
\label{tab:ablation}
\end{table}
\begin{table}[t]
\centering
\renewcommand\tabcolsep{10pt}
\renewcommand\arraystretch{1.2}
\resizebox{\linewidth}{!}{
\begin{tabular}{cccc|ccc}
\hline
\multirow{2}*{MT} &\multirow{2}*{S} &\multirow{2}*{R} &\multirow{2}*{C} &Static &Slow &Fast \\
\cline{5-7}
&&&& \multicolumn{3}{c}{Mean Errors $\downarrow$}\\
\hline
\checkmark & & & &0.0162 &0.4594 &2.9037\\
\checkmark & \checkmark& & &0.0163 &0.4469 &2.8689\\
\checkmark & \checkmark& \checkmark& &0.0179 &0.4579 &4.0773\\
\checkmark & \checkmark& \checkmark& \checkmark &\textbf{0.0161} &\textbf{0.3946} &\textbf{2.7705}\\
\hline
\end{tabular}
}
\caption{Effectiveness of each component in MSRM at 1\% labeled data setting. MT stands for MeanTeacher; S stands for selection; R stands for Re-generate; C stands for Consistency-evaluation.}
\label{tab:MSRM}
\end{table}
\subsection{Main Results}
Table~\ref{tab:main_results} benchmarks results on the nuScenes dataset~\cite{Caesar2019nuScenesAM}. With 1\% labeled data, our SSL method significantly outperforms self-supervised PillarMotion (\eg, 1.9407 vs 3.5504 errors at the fast speed level). With 5\% labeled data, our method surpasses weakly supervised WeakMotionNet. And when with 10\% labeled data, our SSL method can achieve a small performance gap with the fully-supervised baseline, MotionNet (0.0218 vs 0.0256 errors at static level; 0.2746 vs 0.2565 errors at slow level; 1.2030 vs 1.0744 errors at fast level). The results indicate that our method can achieve a good trade-off between annotations and performances.
Additionally, Table~\ref{tab:main_results} also demonstrates the performance improvements gained from unlabeled data using our proposed method. For instance, with our method, the mean errors at fast speed are reduced from 3.7242 to 1.9407 with 1\% labeled data and 99\% unlabeled data; from 2.6661 to 1.4516 with 5\% labeled data and 95\% unlabeled data; from 1.9097 to 1.2030 with 10\% labeled data and 90\% unlabeled data. The qualitative results are shown in Fig.~\ref{fig:qualitative}
\subsection{Ablation Study}
\textbf{Ablation study for the entire framework.} 
In Table~\ref{tab:ablation}, we show the performance with different combinations of Temporal-sampling (TS), MSRM, and BEVMix in the 1\% labeled data setting. Compared to the SSL baseline, Mean-Teacher, TS reduces the mean error by 0.1831 at the fast level, while increasing the error by 0.0376 at the slow level (compare row 1 with row 2). This indicates that TS successfully produces more fast samples for the model to learn. However, it also makes the samples harder to learn. With MSRM, errors are reduced from 0.4594 to 0.3946 at the slow speed level and from 2.9037 to 2.7705 at the fast speed level (compare row 2 with row 3). BEVMix significantly reduces the error from 0.3946 to 0.3497 at the slow level and from 2.7705 to 1.9407 at the fast level (compare row 3 with row 5), demonstrating that BEVMix is an extremely effective data augmentation.
\\
\textbf{Ablation study for the MSRM.} 
The effectiveness of each process in MSRM is shown in Table~\ref{tab:MSRM}. With the only selection, the performance improves slightly (\eg, mean error drop from 2.9037 to 2.8689 at the fast level), demonstrating the effectiveness of discarding unreliable pseudo labels during semi-supervised training. When regenerating pseudo labels without considering local consistency, the prediction results become even worse (\eg, the mean error increases from 2.8689 to 4.0773), which indicates that directly generating from neighbors produces more low-quality pseudo labels. Finally, by selecting reliable pseudo labels again from the re-generated ones, our model achieve better performance (\eg, error drop from 2.9037 to 2.7705). 
\\
\begin{table}[t]
\centering
\renewcommand\tabcolsep{10pt}
\renewcommand\arraystretch{1.2}
\resizebox{\linewidth}{!}{
\begin{tabular}{l|ccc}
\hline
\multirow{2}*{Method} &Static &Slow &Fast \\
\cline{2-4}
& \multicolumn{3}{c}{Mean Errors $\downarrow$}\\
\hline
Mean-Teacher &0.0162 &0.4594 &2.9037\\
+ Mixup &\textbf{0.0106} &0.4282 &2.8902 \\
+ CutMix &0.0166 &0.4510 &2.6157 \\
+ BEVMix &0.0139 &\textbf{0.3581} &\textbf{2.1427} \\
\hline
\end{tabular}
}
\caption{Effectiveness of different mix strategies in 1\% labeled data setting.}
\label{tab:bevmix}
\end{table}
\textbf{Ablation study for the data mix strategy.}
In Table~\ref{tab:bevmix}, we compare our proposed BEVMix with the two most widely used data mixing approaches: Mixup and CutMix. Mixup and CutMix can be considered as randomly replacing cells/areas in the current BEV sequence with cells/areas from the other BEV sequence. With Mixup, static errors are reduced by a large margin, but the fast errors only drop slightly. This is because Mixup can effectively maintain correspondence among static cells throughout temporal windows, but it may easily miss corresponding moving cells.
CutMix has a considerable improvement over the baseline (\eg, mean error drops from 2.9037 to 2.6157 at the fast level). However, our BEVMix can maintain the moving trajectories better and outperforms the Cutmix by a large margin (2.6157 vs 2.1427 error at the fast speed level), showing BEVMix is a more appropriate data mix approach for the class-agnostic motion prediction.
\\
\textbf{Apply TS and BEVMix to fully-supervised training.} We demonstrate that temporal sampling and BEVMix can also provide benefits in the context of fully supervised learning. As illustrated in Table~\ref{tab:fully}, TS and BEVMix can significantly reduce the fast mean error from 0.9878 to 0.8427.
\begin{table}[t]
\renewcommand\arraystretch{0.5}
\centering
\renewcommand\tabcolsep{10pt}
\renewcommand\arraystretch{1.2}
\resizebox{\linewidth}{!}{
\begin{tabular}{l|ccc}
\hline
\multirow{2}*{Method} &Static &Slow &Fast \\
\cline{2-4}
& \multicolumn{3}{c}{Mean Errors $\downarrow$}\\
\hline
MotionNet$^\dag$ &0.0262 &0.2467 &0.9878\\
+ TS &0.0287 &0.2510 &0.9568 \\
+ BEVMix &\textbf{0.0261} &0.2270 &0.8686 \\
+ TS + BEVMix &0.0271 &\textbf{0.2267} &\textbf{0.8427} \\
\hline
\end{tabular}
}
\caption{Effectiveness of TS and BEVMix for fully supervised training. $^\dag$ indicates the results with flip data augmentation.}
\label{tab:fully}
\end{table}
\section{Conclusion}
In this work, we study semi-supervised class-agnostic motion prediction. Specially, we propose two augmentations for the motion prediction task which facilitate the weak-to-strong consistency regularization and significantly improve the performance. Additionally, we propose a novel MSRM module to select and re-generate higher quality pseudo labels, which encourages the model to learn better from the unlabeled data. Experiments show that our semi-supervised method boosts the performance by making use of the unlabeled data and achieving a good trade-off between annotations consumption and performance.

\section{Acknowledgments}
This study is supported under the RIE2020 Industry Alignment Fund – Industry Collaboration Projects (IAF-ICP) Funding Initiative, as well as
cash and in-kind contribution from the industry partner(s).
This research is also supported by the MoE AcRF Tier 2 grant (MOE-T2EP20220-0007).

\bibliography{refs}
\end{document}